\let\oldvec\vec% Store \vec in \oldvec
\documentclass[a4paper]{article}
\let\vec\oldvec% Restore \vec from \oldvec
\usepackage{amsmath, amssymb,amsfonts,amsthm}
\usepackage{mathrsfs}  
\usepackage{graphicx}
\usepackage{xcolor}
\usepackage[linkbordercolor=blue]{hyperref}
\usepackage{booktabs,multirow, rotating}
\usepackage{tabularx}
\usepackage{hyperref}

\usepackage{url}
\urldef{\mailsa}\path|{johannes.berger@iwr.uni-heidelberg.de, schnoerr@math.uni-heidelberg.de}| 

\allowdisplaybreaks[3]
\begin{document}

%%%%%%%%%%%%%%%%%%%%%%%%%%%%%%%%%

%\newtheorem{lemma}[theorem]{Lemma}

%\newtheorem{corollary}[theorem]{Corollary}

%\theoremstyle{definition}
%\newtheorem{definition}{Definition}[section]
%\newtheorem{assumption}{Assumption}[section]
%\theoremstyle{remark}

%\newtheorem{example}{Example}[section]
%\newtheorem{note}{Note}[section]

%\theoremstyle{plain}
%\newtheorem{theorem}{Theorem}[section]
%\newtheorem{proposition}[theorem]{Proposition}

\newtheorem{theorem}{Theorem}
\newtheorem{corollary}[theorem]{Corollary}
\newtheorem{lemma}[theorem]{Lemma}
\theoremstyle{definition}
\newtheorem{definition}[theorem]{Definition}
\newtheorem{example}[theorem]{Example}
\newtheorem{remark}[theorem]{Remark}

%\newtheorem{exercise}{Exercise}
%\newtheorem{remark}{Remark}

%\newtheorem{warning}{Warning}
%%%%%%%%%%%%%%%%%%%%%%%%%%%%%%%%%

\newcommand{\bitem}{\begin{itemize}}
\newcommand{\eitem}{\end{itemize}}
\newcommand{\mc}[1]{\mathcal{#1}}
\newcommand{\mb}[1]{\mathbb{#1}}
\newcommand{\mf}[1]{\mathfrak{#1}}
\newcommand{\ms}[1]{\mathscr{#1}}
\newcommand{\on}[1]{\operatorname{#1}}
\newcommand{\II}{\mathbb{I}}
\newcommand{\N}{\mathbb{N}}
\newcommand{\R}{\mathbb{R}}
\newcommand{\C}{\mathbb{C}}
\newcommand{\F}{\mathcal{F}}
\newcommand{\B}{\mathbb{B}}
\newcommand{\U}{\mathbb{U}}
\newcommand{\EE}{\mathbb{E}}
\newcommand{\V}{\mathbb{V}}
\newcommand{\Q}{\mathbb{Q}}
\newcommand{\Z}{\mathbb{Z}}
\newcommand{\PP}{\mathbb{P}}
\newcommand{\TT}{\mathbb{T}}
\newcommand{\bpm}{\begin{pmatrix}}
\newcommand{\epm}{\end{pmatrix}}
\newcommand{\bsm}{\left(\begin{smallmatrix}}
\newcommand{\esm}{\end{smallmatrix}\right)}
\newcommand{\T}{\top}
\newcommand{\ul}[1]{\underline{#1}}
\newcommand{\ol}[1]{\overline{#1}}
\newcommand{\la}{\langle}
\newcommand{\ra}{\rangle}
\newcommand{\si}{\sigma}
\newcommand{\SI}{\Sigma}
\newcommand{\mrm}[1]{\mathrm{#1}}
\newcommand{\msf}[1]{\mathsf{#1}}
\newcommand{\mfk}[1]{\mathfrak{#1}}
\newcommand{\row}[2]{{#1}_{#2,\bullet}}
\newcommand{\col}[2]{{#1}_{\bullet,#2}}
\newcommand{\df}[2]{\frac{\partial #1}{\partial #2}}
\newcommand{\p}{\partial}
\newcommand{\veps}{\varepsilon}
\newcommand{\toset}{\rightrightarrows}
\newcommand{\w}{\omega}
\newcommand{\gdw}{\Leftrightarrow}
\newcommand{\vphi}{\varphi}
\newcommand{\ora}[1]{\overrightarrow{#1}}
\newcommand{\ola}[1]{\overleftarrow{#1}}
\newcommand{\oset}[2]{\overset{#1}{#2}}
\newcommand{\uset}[2]{\underset{#1}{#2}}
\newcommand{\SE}{\operatorname{SE}_{3}}
\newcommand{\se}{\mathfrak{se}_{3}}
\newcommand{\SO}{\operatorname{SO}_{3}}
\newcommand{\so}{\mathfrak{so}_{3}}
\newcommand{\Ad}{\operatorname{Ad}}
\newcommand{\dist}{\operatorname{dist}}
\newcommand{\etr}{\operatorname{etr}}
\newcommand{\vex}{\operatorname{vex}}
\newcommand{\Psym}{\mathbb{P}_{s}}
\newcommand{\Pskew}{\mathbb{P}_{a}}
\newcommand{\vecso}{\operatorname{vec}_{\mathfrak{so}}}
\newcommand{\vecse}{\operatorname{vec}_{\mathfrak{se}}}
\newcommand{\matso}{\operatorname{mat}_{\mathfrak{so}}}
\newcommand{\matse}{\operatorname{mat}_{\mathfrak{se}}}
\newcommand{\Hess}{\operatorname{Hess}}
\newcommand{\grad}{\operatorname{grad}}
\newcommand{\tr}{\operatorname{tr}}
\newcommand{\argmin}{\operatorname{arg\,min}}
\newcommand{\kronse}{\otimes_{\mathfrak{se}}}
\newcommand{\diag}{\operatorname{diag}}
\newcommand{\Exp}{\operatorname{Exp}}
\newcommand{\D}{\mathbf{D}}
\newcommand{\Id}{\operatorname{Id}}
\newcommand{\G}{\mathcal{G}}
\newcommand{\g}{\mathfrak{g}}
\newcommand{\vecg}{\operatorname{vec}_{\mathfrak{g}}}
\newcommand{\matg}{\operatorname{mat}_{\mathfrak{g}}}
\newcommand{\ad}{\operatorname{ad}}
\newcommand{\Log}{\operatorname{Log}}
\newcommand{\Valuef}{\mathcal{V}}
\newcommand{\Hamilton}{\mathcal{H}}
\newcommand{\Proj}{\mathbf{P}}

\newcommand{\eins}{\mathds{1}}

\newcommand{\cperp}[3]{{#1} \perp\negthickspace\negthinspace\negthickspace\perp {#2} \,|\, {#3}}

\newcommand{\LG}[1]{\mathrm{#1}}
\newcommand{\Lg}[1]{\mathfrak{#1}}

\newcommand{\dom}{\operatorname{dom}}

\newcommand{\st}[1]{{\scriptstyle #1}}
\newcommand{\sst}[1]{{\scriptscriptstyle #1}}

%\mainmatter  % start of an individual contribution

% first the title is needed
\title{Joint Recursive Monocular Filtering \\ of Camera Motion and Disparity Map\thanks{Preprint. The final publication will be available at Springer.}}

%\author{Johannes Berger \and Christoph Schn\"orr}
\author{Johannes Berger\thanks{Support by the German Research Foundation (DFG) is gratefully acknowledged, grant GRK 1653} \\ \href{mailto:johannes.berger@iwr.uni-heidelberg.de}{johannes.berger@iwr.uni-heidelberg.de} 
   \and Christoph Schn\"orr \\ \href{mailto:schnoerr@math.uni-heidelberg.de}{schnoerr@math.uni-heidelberg.de} }

%\affil{Image \& Pattern Analysis Group, Heidelberg University \\ Mathematikon (A), INF 205, 69120 Heidelberg, Germany}

%\email{howard@math.sc.edu}

\maketitle

\begin{abstract}
Monocular scene reconstruction is essential for modern applications such as robotics or autonomous driving. Although stereo methods usually result in better accuracy than monocular methods, they are more expensive and more difficult to calibrate.
In this work, we present a novel second order optimal {\em minimum energy filter} that {\em jointly} estimates the camera motion, the disparity map and also higher order kinematics recursively on a product Lie group containing a novel {\em disparity group}. This mathematical framework enables to cope with non-Euclidean state spaces, non-linear observations and high dimensions which is infeasible for most classical filters. To be robust against outliers, we use a {\em generalized Charbonnier energy function} in this framework rather than a quadratic energy function as proposed in related work. Experiments confirm that our method enables accurate reconstructions on-par with state-of-the-art.
%\keywords{minimum energy filter, monocular reconstruction, camera motion estimation, Lie groups.}
\end{abstract}
\section{Introduction}
\subsection{Overview}
Reconstruction of the scene structure of images and videos is a fundamental building block in computer vision and is required for plenty of applications, e.g. autonomous driving, robot vision and augmented reality. Although stereo methods usually lead to exact reconstruction and work fast, they require calibration of the camera setup and, due to the second camera, these systems are more expensive than single camera systems. Therefore, in this work, we will focus on the monocular approach that consists of reconstructing the scene structure based on the data gained by a {\em single} moving camera. In contrast to the stereo setting, this problem is ill-posed because of the unknown motion parallax. On the other hand, monocular approaches enable cheaper hardware costs.
%can be used as fallback level in a stereo system if one camera fails.

To increase accuracy and robustness of the monocular reconstruction, we want to use temporal information for smoothing and propagation. Thus, we will introduce a mathematical framework based on {\em non-linear} filtering equations which describe the behavior of latent variables and the dependency between latent variables and observations. Since, in this scenario, the state variables, e.g. camera motion, do not evolve on an Euclidean space but a more general Lie group, we cannot use classical filters, such as {\em extended Kalman} filters~\cite{frogerais2012}. Moreover, other state-of-the-art non-linear filters, such as {\em particle filters}~\cite{Doucet2001-CMT}, that can be applied to specific Lie groups \cite{kwon2007}, cannot be easily extended to high dimensional problems~\cite{Daum2003}. Due to these mathematical problems we will use the novel {\em minimum energy filter} on compact Lie groups \cite{saccon2015second} that minimizes a quadratic energy function to penalize deviations of the filtering equations by means of optimal control theory. This filter was shown to be superior to extended Kalman filters on the low dimensional Lie group $\SE$ \cite{Berger-et-al-2015b}. We will demonstrate that this approach can also be successfully applied to high dimensional problems, enabling {\em joint} optimization of camera motion and disparity map. As in \cite{Berger-et-al-2015b}, we will also incorporate higher order kinematics of the camera motion. To be robust against outliers, we will extend the approach of \cite{saccon2015second} from quadratic energy function to a generalized Charbonnier energy function.
 
\subsection{Related Work}
Plenty of methods for depth or disparity map estimation were published during the last decade. We  distinguish between stereo methods (that benefit from the additional information gained from the calibrated camera setup) and monocular methods. Recognized stereo methods include \cite{Hirschmueller2008,psota2015map,vogel20153d} that use the known distance of the cameras (baseline) for accurate triangulation of the scene. These methods also enable reducing the computational effort by using epipolar geometry and by combining local and global optimization schemes. Monocular methods \cite{davison2007monoslam,becker2013variational,engel2013semi,engel2014lsd,pizzoli2014remode,Neufeld-et-al-2015a,bourmaud2015robust} benefit from less calibration effort in comparison to stereo methods, but suffer from a peculiarity of the mathematical setup that prevents to reconstruct the scale of the scene uniquely. 
To increase the robustness and the accuracy of the reconstruction, modern methods incorporate multiple consecutive frames into the optimization procedure. Well-known is bundle adjustment \cite{triggs2000bundle} which optimizes a whole trajectory but cannot be used in online approaches such as sliding window \cite{bellavia2013robust} or filtering methods \cite{becker2013variational,bourmaud2015robust}. 
Filtering methods usually require a suitable modeling of the unknown {\em a posteriori} distribution. However, they suffer from the drawback that the definition of probability densities on non-Euclidean spaces, such as Lie groups, is complicated, although successful strategies to find a solution to this problem have been developed \cite{bourmaud2015continuous,chikuse2012statistics,kwon2007}. Zamani et al. \cite{zamani2012} introduces so-called {\em minimum energy filters} for linear filtering problems for compact Lie groups based on optimal control theory and the recursive filtering principle of Mortensen \cite{Mortensen1968}. This approach was generalized to (non-)compact Lie groups in \cite{saccon2015second} and applied to a {\em non-linear}  filtering problem on $\SE$ for camera motion estimation \cite{berger2015a}.

%\begin{itemize}
%	\item Monocular Reconstruction: \cite{becker2013variational,engel2013semi,engel2014lsd,pizzoli2014remode,Neufeld-et-al-2015a}
%	\item Stereo Reconstrcution: \cite{vogel20153d}
%	\item Minimum Energy Filtering: \cite{Mortensen1968,saccon2015second,zamani2012,Berger-et-al-2015b}
%\end{itemize}

\subsection{Contributions}
Our contributions in this paper add up
\begin{itemize}
	\item to provide a mathematical filtering framework for {\em joint} monocular camera motion and disparity map estimation including higher order kinematics,
	\item to introduce a {\em novel disparity Lie group for inverse depth maps} which avoids additional positive depth constraints such as barrier functions,
	\item to solve the corresponding challenging {\em non-linear} and {\em high-dimensional} filtering problem on a {\em product Lie group} by using novel {\em minimum energy filters},
	\item to provide a {\em generalized Charbonnier} energy function instead of a quadratic energy function \cite{saccon2015second}, which results in robustness against outliers\,.
\end{itemize}

\subsection{Notation}
We use the following spaces: real vector space $\R^{n},$ special orthogonal/Euclidean group $\SO,\SE,$ with their corresponding Lie algebras $\so,\se,$ as well as $\G$ for a general Lie group with Lie algebra $\g.$ Tangent spaces at a point $x$ of $\G$ are denoted by $T_{x}\G.$ A tangent vector $\eta \in T_{x}\G$ can be expressed in terms of a tangent vector $\xi$ on the Lie algebra $\g$ by using the tangent map of the left translation $L_{x}$ evaluated at the identity element $\Id$ of the Lie group, denoted by $\eta =T_{\Id}L_{x}\xi.$ We also use the shorthand $x\xi:=T_{\Id}L_{x}\xi.$ We use the $\ast-$symbol to indicate dual spaces and operators with respect to the Riemannian metric that can be defined by the tangent map as $\la x \eta, x \xi \ra_{x}:= \la \eta, \xi \ra_{\Id}$ for $\eta,\xi \in \g.$ The dual of the tangent map is $T_{\Id}L_{x}^{\ast}\eta=:x^{-1}\eta.$
We denote by $\vecg: \g \rightarrow \R^{n},\matg:\R^{n} \rightarrow \g$ the vectorization and its inverse operation, respectively, where the underlying Lie group $\G$ has dimension $n.$ These operations allow representing the Lie algebra $\g$ in a compact form. $\D f$ denotes the differential of a function, whereas $\D f(x) [\eta]:= \la \D f(x),\eta \ra_{x}$ indicates the directional derivative for a specific direction $\eta$. For compactness, we write $\D_{i} f(x,y,z)$ for the differential of the function $f$ respective the $i-$th component, whereas $\D_{y} f(x,y,z)$ directly addresses a specific variable. $\Hess f [\cdot]$ stands for the Riemanian Hessian on the considered Lie group; the calculation of the latter requires the Riemannian connection $\nabla$ that can be expressed in terms of a connection function $\omega$ on the Lie algebra $\g.$

\section{Model}

In this section we will introduce the mathematical framework of joint monocular camera motion and disparity map estimation from the point of view of (stochastic) filtering. Note, that we will use the notion {\em disparity map} for the inverse of the depth map in this work without using the baseline that is required in stereo settings. In classical filtering theory one wants to determine the most likely state of an unknown process $x=x(t)$ modeled by a perturbed differential equation $\dot{x}(t) = f(x(t)) +\delta(t)$ based on prior perturbed observations $y(s) = h(x(s))+\epsilon(s)$ for $s \leq t,$ which results in a {\em maximum a posteriori} problem. In this work, we require the state space of $x$ to be a Lie group $\G$ which we need to describe non-Euclidean expressions such as camera motions. Using the expressions $\delta = \delta(t)$ and $\epsilon = \epsilon(t)$ to represent model noise and observations noise, respectively, the resulting filtering equations can be written as
\begin{align}
\dot{x}(t) = & x(t)\bigl( f(x(t)) + \delta(t) \bigr)\,, \quad x(t_{0}) = x_{0} \,, \label{eq:state}\\
y(t) = & h(x(t)) + \epsilon(t)\,. \label{eq:observation}
\end{align}
The state equation \eqref{eq:state} is modeled on a Lie Group $\G$ by means of the tangent map of the left translation at identity and functions $f,\delta \in \g$ such that $\dot{x}(t)\in T_{x}\G.$ In the following sections we will introduce the state space of $x,$ the propagation functions $f$ and the observation function $h.$
\subsection{State Space}
The camera motion is modeled on the Special Euclidean group $\SE:= \{\bsm R & w \\ 0 & 1\esm \vert R \in \SO, w \in \R^{3} \},$ and we also use a higher order kinematics (e.g. acceleration of camera) modeled by a vector $v \in \R^{6}$. The disparity map can be represented by a large vector $d_{i} \in \R^{\vert \Omega  \vert},$ resulting in an own dimension for each pixel in the image. However, the depth must always be positive and we want to avoid additional constraints within our optimization. Therefore, we introduce a novel Lie group for the inverse of the depth, denoted by $(0,1)^{\vert \Omega \vert}$ which is defined as follows:
\begin{definition}[Lie group $(0,1)^{n}$ (Disparity group)]\label{def:inverse-lie-group}
By denoting $d_{i}(z,t):=\tfrac{1}{d(z,t)}\in (0,1)$ the inverse of the depth we define the Lie group $(0,1)^{n}$ with group action for $x, y \in (0,1)^{n}$ as
\begin{align*}
x \circ y \mapsto & \bigl((x^{-1} - \mathbf{1}) \cdot (y^{-1}-\mathbf{1}) + \mathbf{1}\bigr)^{-1} =  \frac{xy}{\mathbf{1}-x-y+2xy}\,.
% = & \Bigl( \frac{(1-x)(1-y)+xy}{xy} \Bigr)^{-1} \\
\end{align*}
The (Lie group inverse) can be computed as $i(x):= \mathbf{1}-x.$ This results in the identity element $\Id = \tfrac{\mathbf{1}}{\mathbf{2}},$ i.e. a vector full of $1/2.$ The exponential map $\Exp_{(0,1)^{n}}: \R^{n} \rightarrow (0,1)^{n}$ and the logarithmic map $\Log_{(0,1)^{n}}:\R^{n}\rightarrow (0,1)^{n}$ are given through $ x \mapsto\frac{e^{\mathbf{4}x}}{\mathbf{1} + e^{\mathbf{4}x}}$ and $x \mapsto \tfrac{1}{4}\log\Bigl(\frac{x}{\mathbf{1}-x} \Bigr),$ respectively. All operations apply component-wise to the vectors involved.
\end{definition}
Using $\SE$ for the camera motion, $\R^{6}$ for the acceleration of the camera and the Lie group given through definition \ref{def:inverse-lie-group} for the disparity map, we find the product Lie group $
\G$ for our state space, i.e.
\begin{equation} \label{eq:def-LieGroupG-depth}
\G := \SE \times \R^{6} \times (0,1)^{\lvert \Omega \rvert} \,.
\end{equation}
%For the filtering problem, we will require some prior knowledge on the state variable $x =(E,v,d_{i}) \in \G,$ which is expressed in terms of an ordinary differential equation or a corresponding discrete propagation.
\subsection{Propagation of the Camera Motion}
%We assume a that a specific moment $m$ of the camera motion is constant such that camera motion can be expressed as $m-$th order dynamical system cf. \cite{Berger-et-al-2015b}.
For propagation of the camera we will use a second order kinematic model that can be expressed as second order differential equation on $\SE$ as in \cite{Berger-et-al-2015b}, which is
\begin{align}
\begin{split}\label{eq:propagation-E}
\dot{E}(t) = & E(t) \matse(v(t)), \quad E(t_{0}) = E_{0}\,, \\
\dot{v}(t) = &  \mathbf{0}, \quad v(t_{0}) = v_{0}\,, \\
\end{split}
\end{align}
where $E=E(t) \in \SE$ and $v=v(t) \in \R^{6}.$  
\begin{remark}
Since $E$ describes the local camera motion from frame to frame, a first order model $\dot{E}(t)=\mathbf{0}$ corresponds to a constantly moving camera, i.e. with constant velocity. Thus, the model \eqref{eq:propagation-E} describes a constant acceleration in the global camera frame.
\end{remark}

\subsection{Discrete Propagation of the Disparity Map}\label{sec:discrete-propagation-disparity}
\begin{figure}[htbp]
\begin{center}
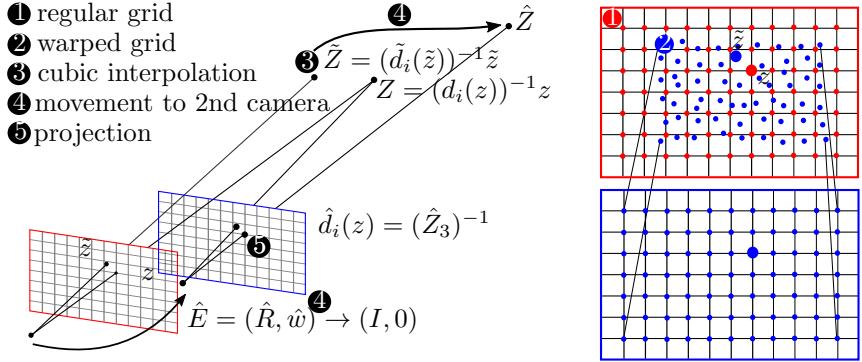
\caption{Discrete propagation of the disparity map}
\label{fig:discrete-prop}
\end{center}
\end{figure}

The propagation consists of mapping the image grid forward by an estimate of the motion $\hat{E} = (\hat{R}, \hat{w}),$ by cubic interpolation of the depth on the irregular grid and back-projection of the resulting scene points. This leads to the following algorithm, which is also depicted in Figure \ref{fig:discrete-prop}. 
\begin{enumerate}
	\item Start with the disparity map $d_{i}$ on regular image grid in camera $(I,0)$.
	\item Warp the image grid forward into next image (camera estimate $\hat{E}(t)$) by using current disparity map $d_{i}$ to get a grid with points $\tilde{z} = \pi(\hat{R} \bsm z \\1 \esm (d_{i}(z))^{-1} +  \hat{w}),$ where $\pi: \R^{3} \rightarrow \R^{2}$ is given through $(z_{1},z_{2},z_{3})^{\T}\mapsto (z_{3})^{-1} (z_{1},z_{2})^{\T}$.
	\item Perform cubic interpolation on the warped grid $\tilde{z}$ given the values $(z,d_{i}(z))$ which gives the new depth map $(\tilde{z}, \tilde{d}_{i}(\tilde{z}))$ in frame $(I,0)$.	
	\item Move $\tilde{Z} =\bsm \tilde{z} \\1 \esm (\tilde{d_{i}}(\tilde{z}))^{-1}$ to second camera to obtain $\hat{Z} = \hat{R}^{\T} \bigl( \tilde{Z} - \hat{w} \bigr)$.
	\item Recognize the propagated disparity map as third component, $\hat{d_{i}}(z) = (\hat{Z}_{3})^{-1}$.
\end{enumerate}
\subsection{Camera Motion and Disparity Map induced Optical Flow}
Since the state space $\G$ consists of the camera motion $E(t)$ and the disparity map $d_{i}(\cdot,t)$ (inverse of depth map), we require observations that depend on both variables. It is well-known that from a given disparity map and a given camera motion the correspondences between a pair of consecutive images expressed as {\em optical flow} can be uniquely determined if the scene is static. To be precise, the dependency between the optical flow vector $u(z,t)$ at a position $z \in \Omega$ can be expressed with the following {\em non-linear} relation, where we denote by $R(t)\in \SO$ and $w(t)\in \R^{3}$ the rotational and translational component of the camera motion $E(t)=(R(t),w(t))\in\SE,$ respectively. For details see \cite[Eq.\ (6)]{becker2013variational}.
\begin{equation}\label{eq:egomotion-depth-induced-OF}
u(z,t) + z = \pi\Bigl( R(t)  \bsm z \\ 1 \esm (d_{i}(z,t))^{-1} + w(t) \Bigr).
\end{equation}

\subsection{Overall Filtering Model}
The function $f(x(t)): \G \rightarrow \mf{g}$ in \eqref{eq:state} can now be defined as follows:
\begin{equation} \label{eq:def-f-depth}
f(x(t)) := (f_{E}(x(t)), f_{v}(x(t)), f_{d_{i}}(x(t)))\,,
\end{equation}
with component functions $f_{E}(x(t)):= \matse(v(t)),$ and  $f_{v}(x(t)) := \mathbf{0_{6}}$ as in \eqref{eq:propagation-E} as well as $f_{d_{i}}(x(t)) := \mathbf{0}_{\lvert \Omega \rvert}.$ Beside this continuous propagation step we also incorporate discrete updates of the disparities as described in section \ref{sec:discrete-propagation-disparity}. 

By setting $y_{z}(t):=u(z,t)-z$ and $h_{z}:\G\rightarrow \R^{2},$ as the right hand side of \eqref{eq:egomotion-depth-induced-OF}, we find the following observation equations by adding noise $\epsilon_{z}(t) \in \R^{2}$ for all $z \in \Omega.$
\begin{equation}\label{eq:observation-equation-depth}
y_{z}(t) = h_{z}(x(t)) + \epsilon_{z}(t)\,, \qquad  z \in \Omega\,.
\end{equation}

\subsection{Objective Function}
Minimum energy filtering requires to define an energy function that penalizes the model and observation noise. In contrast to \cite{saccon2013second}, that we will follow in this work, we will not use quadratic energy functions but an energy function that is a smooth approximation of the $L^{1}-$norm. The reason is that we want to reduce the influence of outliers in the observations that may cause numerical problems because the gradient grows linearly.  The norm of the gradient of the proposed $L^{1}$ penalty function is bounded. A smooth approximation to the non-differentiable $L^{1}-$norm is the generalized charbonnier penalty function that is smooth ($C^{\infty}$) and has linear growth, such that we use it for $\phi,$ i.e. $\phi(x):= (x  + \nu)^{\beta} - \nu^{\beta}.$ With this notation and the shorthand $\lVert x \rVert_{Q}^{2}:=x^{\T}Qx$  the energy function reads
\begin{align}\label{eq:energy-depth-regularizer-L1}
\begin{split}
\mc{J}(\delta,&  \epsilon, x ;t):=  \tfrac{1}{2}\lVert x-x_{0} \rVert_{R_{0}^{-1}}^{2}+ \int_{t_{0}}^{t}  \Bigl( \tfrac{1}{2} \lVert \vecg(\delta(\tau)) \rVert_{R^{-1}}^{2}+ \sum_{z\in \Omega} \phi(\tfrac{1}{2}\lVert \epsilon_{z}(\tau)\rVert_{Q_{z}^{-1}}^{2}\bigr)\Bigr)d\tau, 
\end{split}
\end{align}
where $Q_{z},R_{0}$ and $R$ are symmetric and positive definite matrices.

\subsection{Optimal Control Problem}
After replacing the observation noise $\epsilon_{z}(t)$ by the residual $\epsilon_{z}(t)=\epsilon_{z}(x(t),t):=y_{z}(t) - h_{z}(x,t)$ in \eqref{eq:energy-depth-regularizer-L1} we want to minimize the energy function $\mc{J}(\delta, x,x(t_{0});t)=\mc{J}(\delta, \epsilon(x),x(t_{0});t)$ regarding the model noise $\delta(t)$ with respect to the differential equation \eqref{eq:state} yielding the {\em value function}
\begin{equation}\label{eq:ValueFunction}
\Valuef(x(t),t,x(t_{0})):=  \min_{\delta\vert_{[t_{0},t]}} \mc{J}(\delta,x;t) \quad \mbox{subject to \eqref{eq:state}}.
\end{equation}
Calculation of the value function requires to introduce the time-varying (left-trivialized) Hamiltonian function $\tilde{\Hamilton}: \G \times \g^{\ast} \times \mf{g} \times \R \rightarrow \R$ that is given through
\begin{align}
\begin{split}
\tilde{\Hamilton}(x, \mu, \delta, t):=  & \bigl(\tfrac{1}{2}  \lVert \vecg(\delta(t))\rVert_{R^{-1}}^{2} + \sum_{z \in \Omega}\phi(\tfrac{1}{2}\lVert y_{z}(t)-h_{z}(x(t)))\rVert_{Q_{z}}^{2} \bigr) \\
& \hspace{3cm} - \la \mu, f(x(t))+\delta(t) \ra_{\Id}\,.
\end{split}
\end{align}
Owing to the {\em Pontryagin minimum principle} \cite{pontryagin1962ef} we find the minimizing argument of the value function \eqref{eq:ValueFunction} by minimizing the Hamiltonian $\tilde{\Hamilton}$ with respect to $\delta.$ Since the Hamiltonian is convex in $\delta$ we obtain a unique minimum $\delta^{\ast}=\matg\big(R \vecg(\mu)   \bigr)$ resulting in the optimal Hamiltonian $\Hamilton(x,\mu, t): \G \times \g^{\ast} \times \R \rightarrow \R$ given through $\Hamilton(x,\mu,t):= \tilde{\Hamilton}(x,\mu,\delta^{\ast}, t)$ such that
\begin{align*}
\Hamilton(x,\mu,t)=& - \la \mu,  f(x(t) \ra_{\Id} - \tfrac{1}{2}\lVert  \vecg(\mu) \rVert_{R}^{2} + \sum_{z \in \Omega}\bigl( \phi \bigl( \tfrac{1}{2}\lVert y_{z}(t)-h_{z}(x(t)) \rVert_{Q_{z}}^{2}\bigr)  \bigr)\,.
\end{align*}
In the case of a linear-quadratic control problem this optimal Hamiltonian satisfies the (left-trivialized) Hamilton-Jacobi-Bellman equation, i.e. 
\begin{equation}\label{eq:HJB-equation}
\frac{\p}{\p t} \Valuef(x,t) - \Hamilton(x,x^{-1}\D_{1}\Valuef(x,t),t) = 0\,.
\end{equation}
Here, $\D_{1} \Valuef(x,t)\in T_{x}^{\ast}\G$ is an element of the cotangent space.

\begin{remark}
Note that our control problem has neither {\em linear control dynamics} nor a {\em quadratic} energy function. Thus, we have no guarantee that the HJB equation is a necessary {\em and sufficient} condition for optimality. Instead we require a good initialization to gain an optimal reconstruction. However, we will show that a fairly general initialization will lead to good reconstructions.
\end{remark}
\subsection{Recursive Filtering Principle and Truncation}
Computation of the total time derivative of the necessary condition 
$$\D_{1} \Valuef(x,t,x(t_{0}))=\mathbf{0}\,,$$
and insertion of the HJB equation \eqref{eq:HJB-equation} leads to the following lemma that gives a recursive description of the optimal state $x^{\ast} = x^{\ast}(t)$ (cf. \cite[Eq. (37)]{saccon2015second}).
\begin{lemma}\label{lem:lem1} The evolution equation of the optimal $x^{\ast}$ state is given through
\begin{equation}\label{eq:evolution-eq-x-depth}
\dot{x}^{\ast}(t) = x(t)\Bigl(f(x^{\ast}(t)) - \hat{Z}(x^{\ast}(t),t)^{-1}\circ x^{-1}(\D_{1}\Hamilton(x^{\ast}(t),\mathbf{0},t))  \Bigr)\,,
\end{equation}
where $\hat{Z}: \mf{g} \rightarrow \mf{g}^{\ast}$ is the left-trivialized Hessian of the value function given through
\begin{equation}\label{eq:def-Z-depth}
\hat{Z}(x^{\ast},t) \circ \eta= (x^{\ast})^{-1} \Hess \Valuef(x^{\ast}(t),t,x(t_{0}))[x^{\ast} \eta]\,, \quad \eta \in \mf{g}\,.
\end{equation}
\end{lemma}
Because the non-linear filtering problem is infinite dimensional we will replace the exact operator $\hat{Z}$ by an approximation $Z: \g \rightarrow \g^{\ast}$ which can be obtained by truncation of the full evolution equation of $Z$. But still the operator  $Z(x^{\ast},t)$ on $\g$ is complicated such that we introduce a matrix representation $P(t)$ that is defined through the relation $\vecg(Z(x^{\ast},t)^{-1} \circ \eta)=:P(t)\vecg(\eta).$

\begin{lemma} \label{lem:lem2}The matrix representation of the approximation of the operator $\hat{Z}$ evolves regarding the following matrix Riccati equation
\begin{align}
\dot{P}(t) = R + C(x^{\ast},t)P(t) + P(t)C(x^{\ast},t)^{\T} - P(t) H(x^{\ast},t) P(t),
\end{align}
where the matrix $R$ is the weighting matrix in the energy function \eqref{eq:energy-depth-regularizer-L1} and the matrices $C$ and $H$ are given for $\eta \in \g$ through
\begin{align*}
C(x^{\ast},t)P(t)\vecg(\eta) := & \vecg\bigl((x^{\ast})^{-1}\D_{2}(\D_{1} \Hamilton(x^{\ast},\mathbf{0},t))[Z(x^{\ast},t)\circ \eta] \bigr) \\ 
& \hspace{-2cm} + \vecg( \omega_{\D_{2} \Hamilton(x^{\ast},\mathbf{0},t) }^{\leftrightharpoons \ast} \circ Z(x^{\ast},t) \circ \eta  ) + \vecg(\omega_{(x^{\ast})^{-1}\dot{x^{\ast}}}^{\ast} \circ Z(x^{\ast},t)\circ \eta)\,,  \\
H(x^{\ast},t)\vecg(\eta):= & \vecg((x^{\ast})^{-1} \Hess_{1} \Hamilton(x^{\ast},\mathbf{0},t)[x\eta]   )\,.
\end{align*}
Here, $x\omega_{\xi}\eta := \nabla_{x \xi} x \eta$ denotes the connection function on the Lie algebra $\g$ of the Levi-Civita connection $\nabla_{\cdot}\cdot$ for $\xi, \eta \in \g$ and $x \in \G$, and $\omega_{\xi}^{\leftrightharpoons \ast}$ is the dual of the ``swaped'' connection function $\omega_{\xi}^{\leftrightharpoons}\eta:= \omega_{\eta}\xi$ (cf. \cite{saccon2013second}).
\end{lemma}  

By insertion of the expression $P$ into \eqref{eq:evolution-eq-x-depth} and by evaluation of the expressions in Lemma \ref{lem:lem1} and \ref{lem:lem2} we obtain the final minimum energy filter that consists of continuous propagation of the states with a discrete update of the disparity map.

\begin{theorem}
The second order minimum energy filter with additional discrete propagation step for the disparity map is given through the following evolution equations of the optimal state $x^{\ast} \in \G$ as well as the second order operator $P \in \R^{(12 + \lvert\Omega\rvert) \times (12 + \lvert \Omega \rvert)}.$
\begin{align}
\dot{x}^{\ast}(t) = & x^{\ast}(t) \bigl(f(x^{\ast}(t)) -\matg(P(t)\vecg(G(x^{\ast}(t),t) ))\bigr), \, \label{eq:diff-x-ast} \\
\dot{P}(t) = & R + C(x^{\ast},t)P(t) + P(t)C(x^{\ast},t)^{\T} - P(t) H(x^{\ast},t) P(t), \, \label{eq:diff-P}
\end{align}
with initial conditions $x^{\ast}(t_{0})=x_{0}$ and $P(t_{0})=R_{0},$ where $R_{0}$ is the matrix in \eqref{eq:energy-depth-regularizer-L1}. $G(x^{\ast},t)=(G_{E}(x^{\ast}), \mathbf{0},G_{d_{i}}(x^{\ast})) \in \g$ denotes the Riemannian gradient of the Hamiltonian in \eqref{eq:evolution-eq-x-depth} with components $G_{E}$ and $G_{d_{i}}.$ 
\end{theorem}
The numerical integration of these equations between the time steps $t_{k-1}$ and $t_{k}$ correspond to the update step of a filter, where the updates are assumed to be piecewise constant. After each update step the disparity map is propagated forward using the procedure in Fig.~\ref{fig:discrete-prop} that result in the final filter.
\begin{remark}
The expressions for $C(x^{\ast},t), H(x^{\ast},t) $ and $G(x^{\ast},t)$ can be calculated explicitly but require 
matrix calculus and differential geometry. The resulting expressions become involved such that we refer the interested reader to the supplemental material\footnote{\url{http://hciweb.iwr.uni-heidelberg.de/people/johannesberger}}.
\end{remark}
\begin{remark}
The optimal state can be calculated by geometric numerical integration of the ordinary differential equations \eqref{eq:diff-x-ast} and \eqref{eq:diff-P}, e.g. Crouch-Grossman methods (cf. \cite{hairer2006geometric}). During numerical integration it is important to keep the matrix $P$ sparse, therefore we  set the off-diagonal entries of the lower right part of $P$ (that addresses the disparities) after each iteration to zero.
\end{remark}
\section{Experiments}
\textbf{Preprocessing} As stated above, our method requires precise optical flow as input. Since we propose a monocular method we also demand that the optical flow is computed from two consecutive image frames without stereo information. For this reason we used the well-known {\em EpicFlow} approach \cite{Revaud2015}. The  matches are computed with {\em Deep Matching} \cite{weinzaepfel2013deepflow}; the required edges are from \cite{PMT}. 

\textbf{Choice of the weighting matrices} Monocular methods suffer from the fact that observations that appear close to the epipole (focus of expansion) are orthogonal to the camera motion such that these regions cannot be reconstructed correctly. Therefore we use the weighting term from \cite[Eq.~(14)]{becker2013variational} for the weighting matrix $Q$ that decreases the influence of the data term in regions close to the epipole.
%\paragraph{Choice of the penalty function} We decided to use the Charbonnier penalty function for the data term to be robust against outliers and to enable stable numerical integration on the Lie group. A quadratic energy function leads to inaccuracies during reconstruction since the gradients get to large as shown in Fig.~\textbf{refXXX}

\textbf{Outlier detection} To remove outliers, we computed the backward flow from frame $i$ to $i+1$ as well as the forward flow from frame $i+1$ to $i.$ In regions where these flows are not consistent with each other, we decreased the weight of the term $R$ such that the filter has less ability to fit to the data and the discrete disparity map propagation from section \ref{sec:discrete-propagation-disparity} reduces the error. 

\textbf{Scale correction} As monocular approaches cannot estimate the scale of a scene without prior knowledge about invariants in the scene, we corrected the scale by calculating of the pixel-wise quotient of the disparities and taking its median as scale $s: = \on{median}\{ d_{i}^{\mbox{gt}}(z,t)/d_{i}^{\mbox{est}}(z,t)\vert z \in \Omega^{\ast}\},$ where $\Omega^{\ast}$ denotes the image domain without points which are close to the epipole  ($<50$ pixel distance).
%\paragraph{Estimation of scale parameter} For evaluation of the disparity maps from \cite{becker2013variational} and our approach we require to correct scale of the scene that {\em a priori} cannot be determined from the image data itself but additional information about invariants in the scene, e.g. the camera height (cf. \cite{Kitt20116872}). On the KITTI benchmark \cite{Geiger2012CVPR} we selected a fixed region of interest $\Omega_{ROI}$ on the street  to find the plane normal $\lambda \in \R^{3}$ by minimizing the energy function 
%$\mathcal{J}_{nor}(\lambda) = \sum_{z\in \Omega_{ROI}} \lVert \pi( R^{\T}(I_{3} -w \lambda^{\T} ) \bsm z \\ 1 \esm ) -  z \rVert_{2}^{2},$ cf. \cite{Neufeld-et-al-2015a}. The camera height assumed to be the length of $\lambda,$ i.e. $h_{est} =\lVert \lambda \rVert_{2};$ the scale parameter $s \in \R_{>0}$ can be determined by the quotient of estimated $h_{est}$ and true camera height $h_{gt},$ i.e.~$s = h_{est} / h_{gt}.$ 
%\vspace{-0.3cm}
\subsection{Qualitative Results}
\begin{figure}[tb]
\begin{center}
\begin{tabular}{ccc}
\includegraphics[scale=0.09]{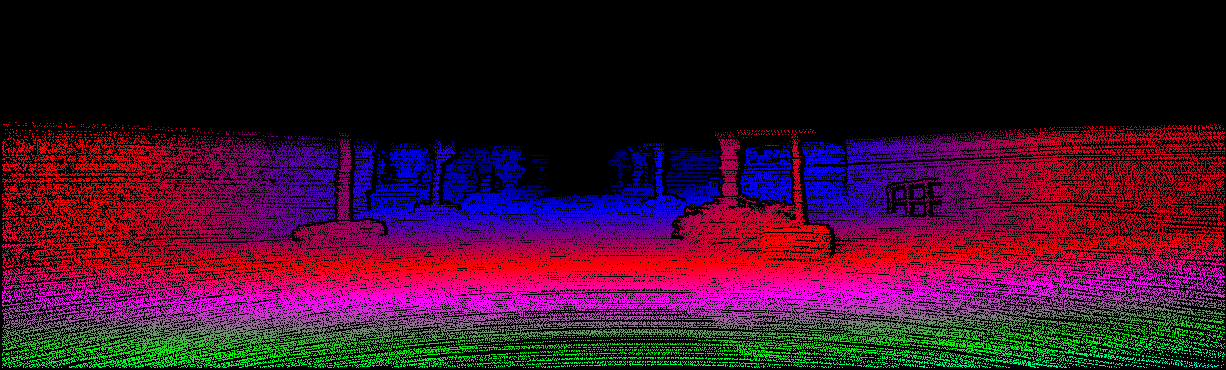} & \includegraphics[scale=0.09]{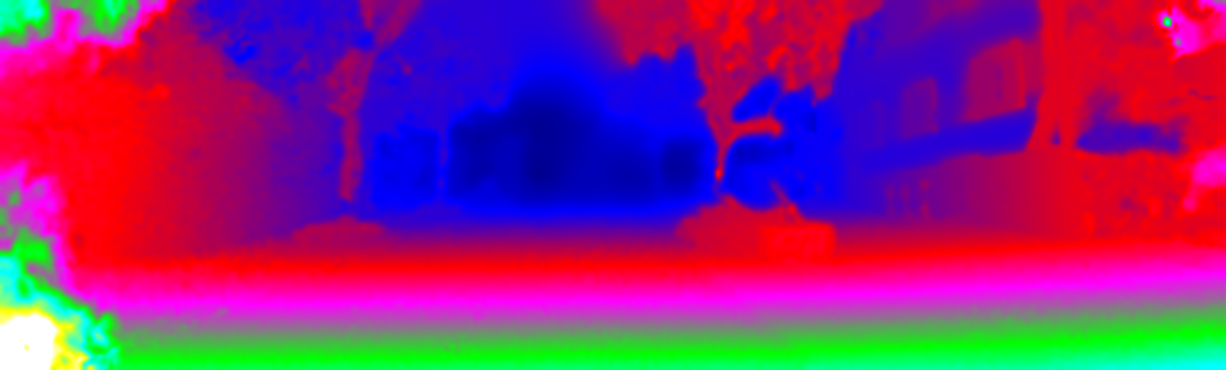} & \includegraphics[scale=0.09]{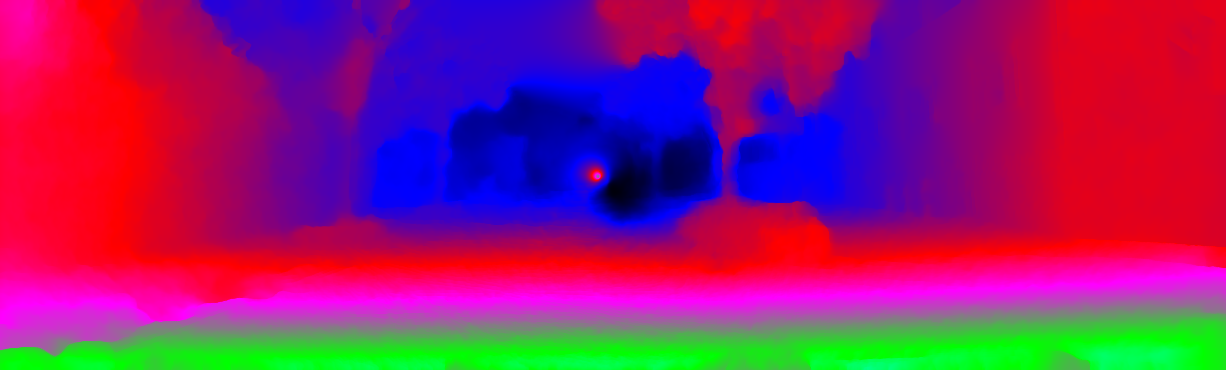} \\
\includegraphics[scale=0.09]{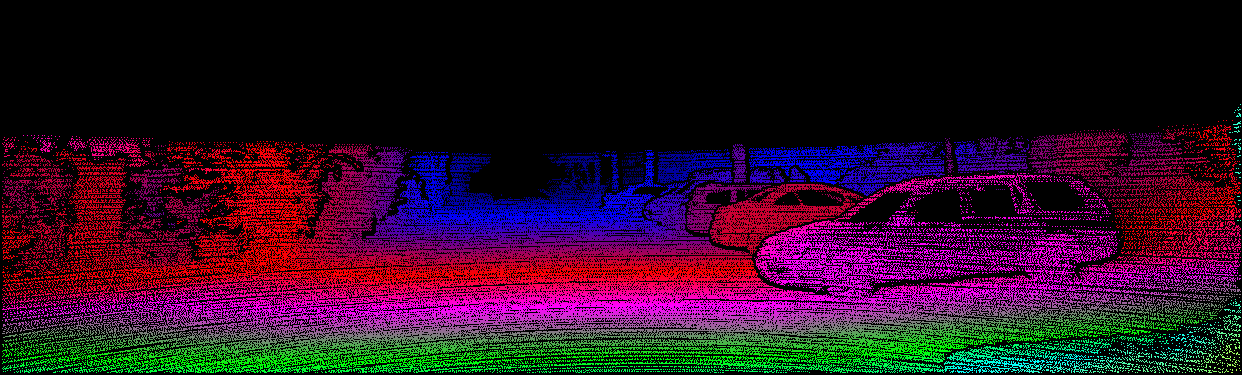} & \includegraphics[scale=0.09]{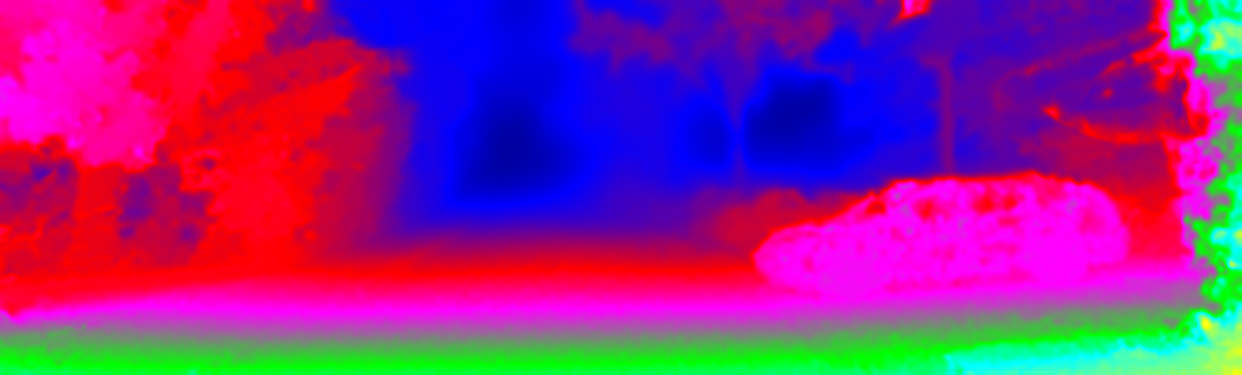} & \includegraphics[scale=0.09]{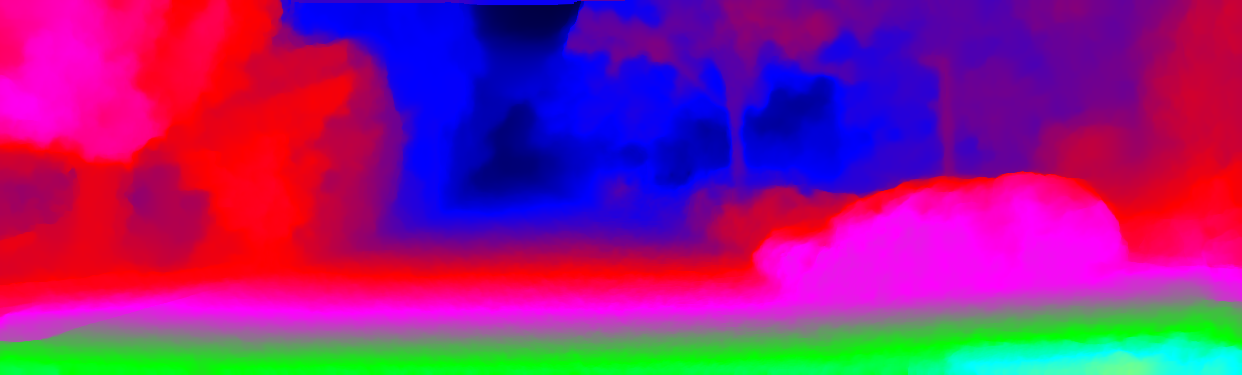} \\
\includegraphics[scale=0.09]{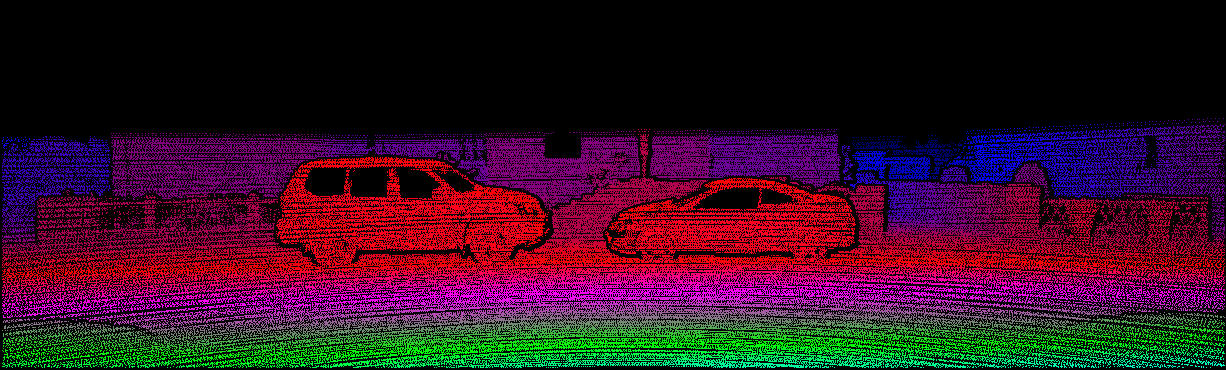} & \includegraphics[scale=0.09]{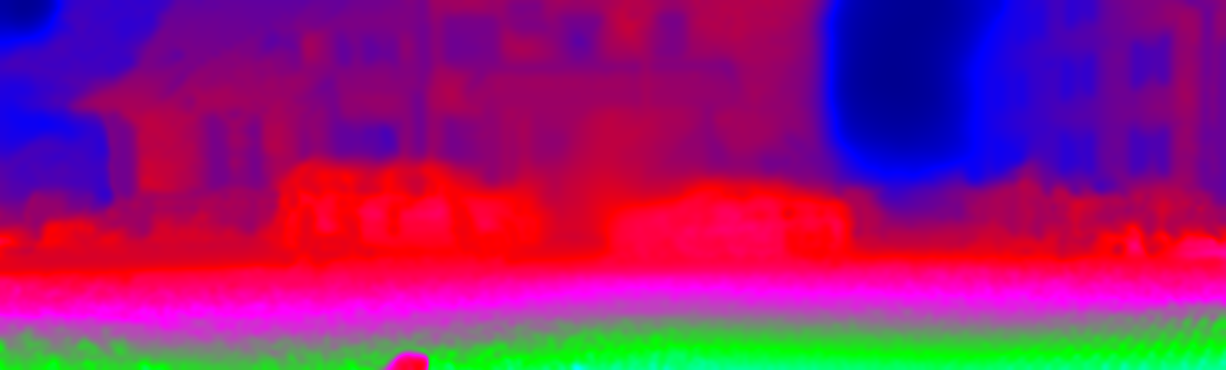} & \includegraphics[scale=0.09]{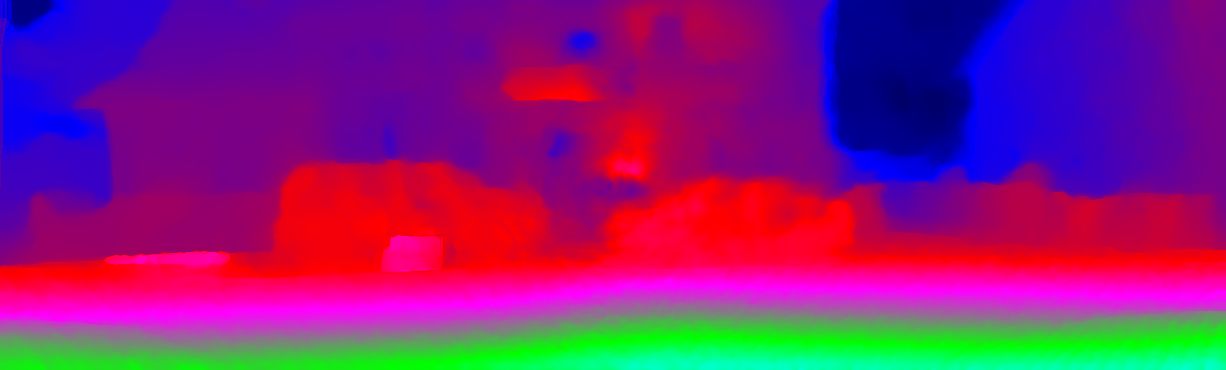} \\
\includegraphics[scale=0.09]{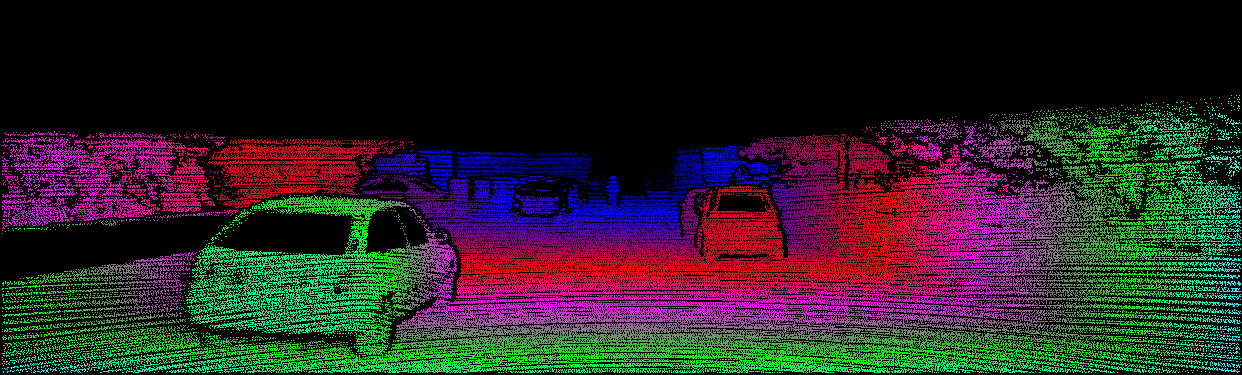} & \includegraphics[scale=0.09]{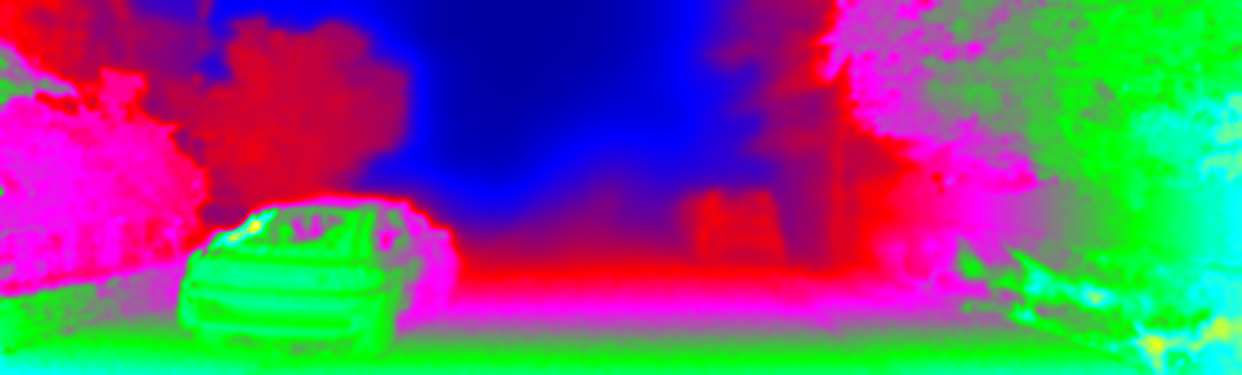} & \includegraphics[scale=0.09]{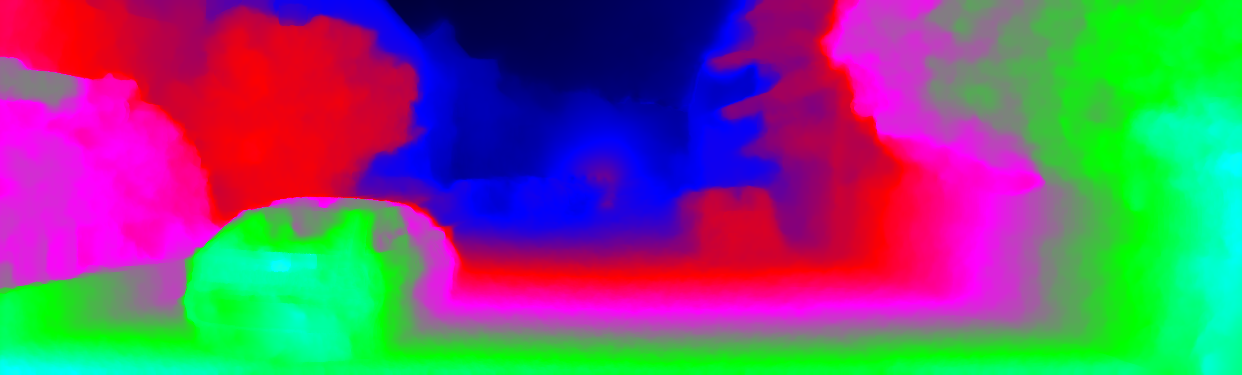} \\
\includegraphics[scale=0.09]{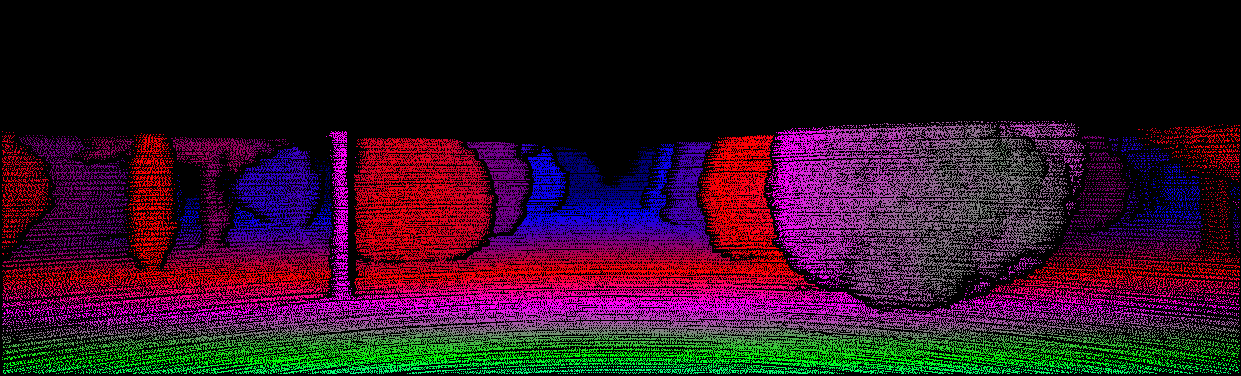} & \includegraphics[scale=0.09]{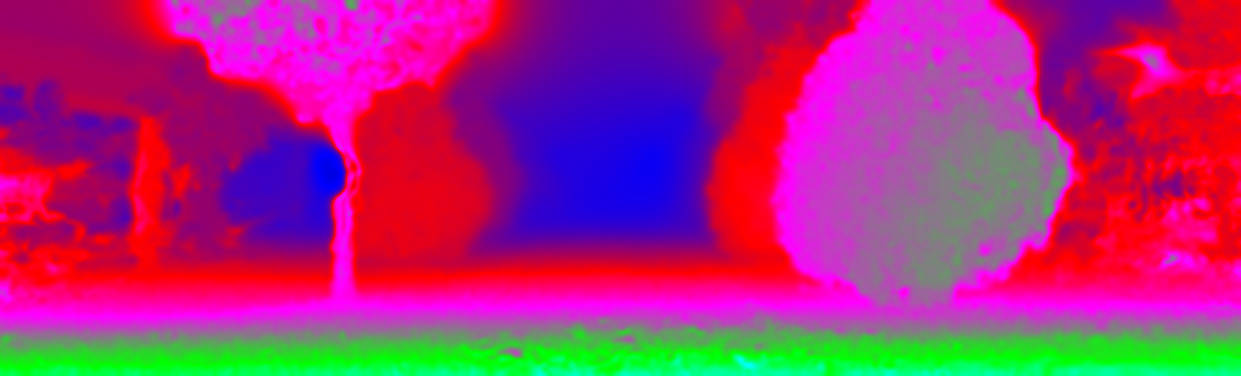} & \includegraphics[scale=0.09]{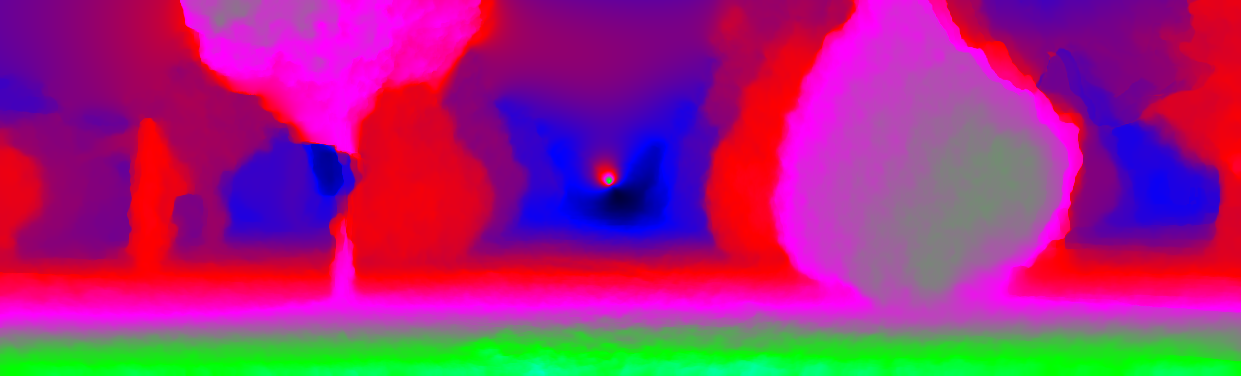} \\
\includegraphics[scale=0.09]{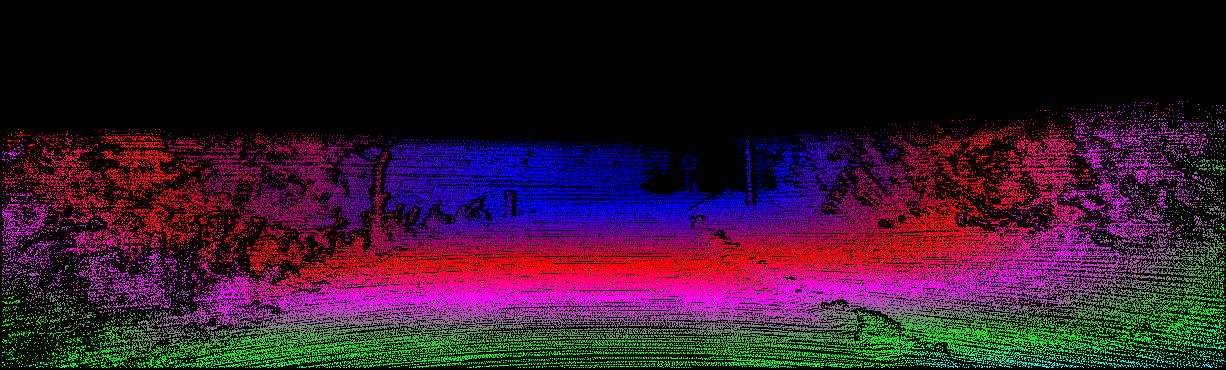} & \includegraphics[scale=0.09]{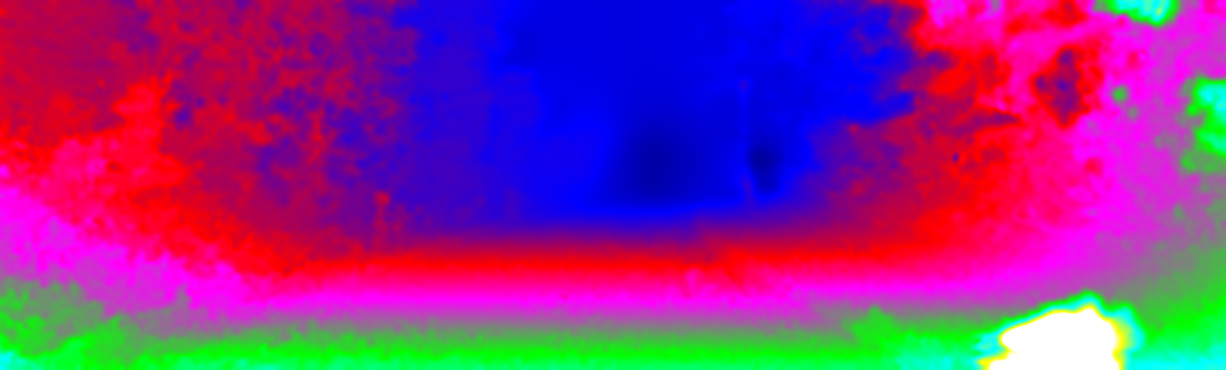} & \includegraphics[scale=0.09]{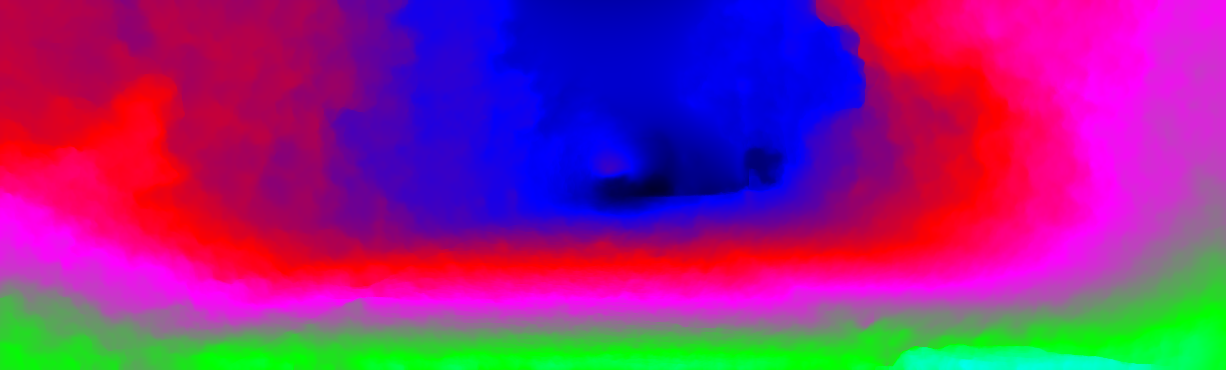} \\
\includegraphics[scale=0.09]{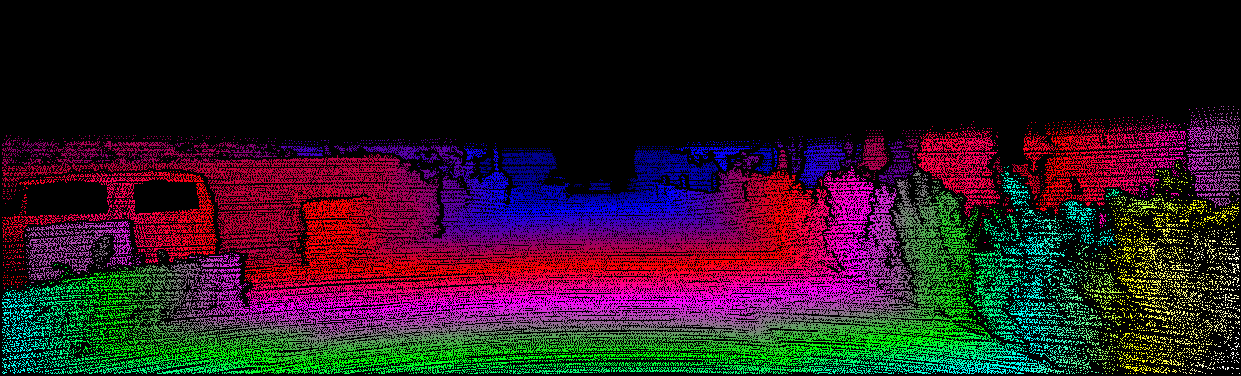} & \includegraphics[scale=0.09]{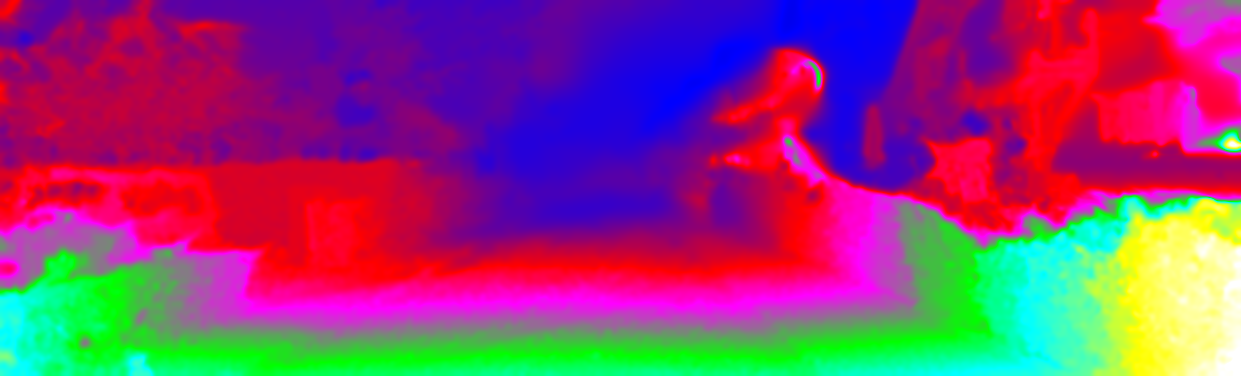} & \includegraphics[scale=0.09]{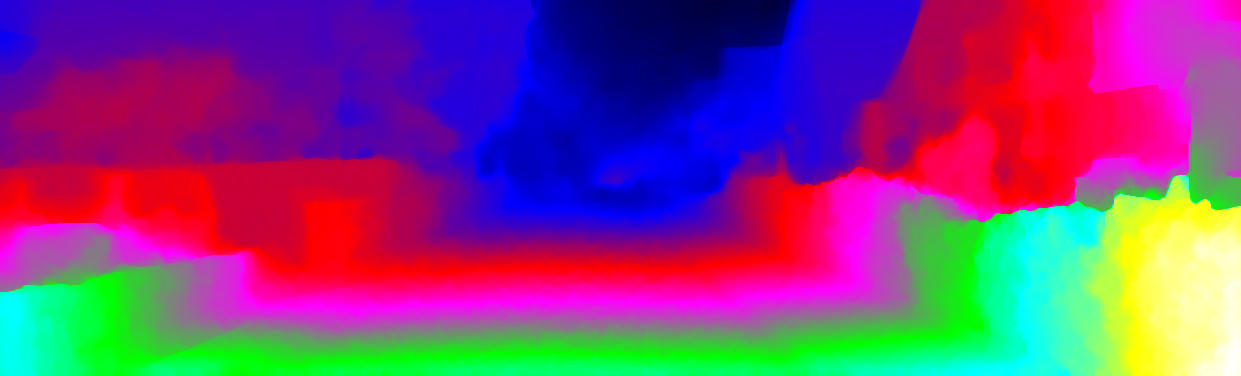} \\
\end{tabular}
\caption{{\em Best viewed in color.} Reconstruction of the disparity maps; left column: ground truth  from the KITTI stereo benchmark, middle column: monocular method of Becker et al. \cite{becker2013variational}, right column: reconstruction with our monocular method. Although in the quantitative evaluation both methods perform equally, one can recognize that our method results in sharper corners. Due to spatial regularization \cite{becker2013variational} reconstructs regions close to the epipole better.}
\label{fig:comparison_disparities}
\end{center}
\end{figure}

In Fig.~\ref{fig:comparison_disparities} we compared the reconstruction of the disparity map of our method with the results from \cite{becker2013variational} and the ground truth. One can recognize that our method preserves small details and depth discontinuities better than \cite{becker2013variational} and returns sharper edges.
\subsection{Quantitative Results}
We evaluated the mean amount of pixels in $\Omega^{\ast}$ with a disparity error larger than three pixel for both occluded and not occluded scenarios in Table~\ref{tab:quant-disparities}. We are slightly inferior towards Becker et al. \cite{becker2013variational}; however, unlike \cite{becker2013variational} we do not have spatial regularization within our optimization which explains the differences.
\begin{table}[htbp]
\caption{Evaluation of the mean disparity errors.}
\begin{center} % @{\extracolsep{\fill}}
\begin{tabularx}{\columnwidth}{l @{\extracolsep{\fill}} r r r r r r} \toprule
			&  $p_{3px} [\%]$ (occ) 	   & 	$p_{5px} [\%]$ (occ) &  $p_{3px} [\%]$ (noc)  & 	$p_{5px} [\%]$ (noc)  \\ \midrule
Becker et al. \cite{becker2013variational}	& 17.74	 & 10.82  &  17.63     & 10.72 \\ 
our approach  &  19.24	& 10.69     &  19.14       & 10.59 \\  \bottomrule
\end{tabularx}
\end{center}
\label{tab:quant-disparities}
\end{table}

\section{Conclusion}
We provided a sound mathematical filtering framework for monocular scene reconstruction based on novel minimum energy filters, extending the classical quadratic energy function from Saccon et al.~\cite{saccon2015second} to a generalized Charbonnier energy function. We demonstrated that the proposed filter copes with challenging mathematical issues, such as a non-Euclidean state space, non-linear filtering equations based on projections, as well as high dimensions; in fact, these difficulties are infeasible for most classical stochastic filters. The introduced {\em disparity group} enables filtering without additional constraints making the model relatively compact. Our experiments confirmed that the proposed filter is almost as accurate as other state-of-the-art monocular and recursive methods without having an own regularization within the model.

\bibliographystyle{plain}     % basic style, author-year citations
\bibliography{main}   % name your BibTeX data base

\end{document}